\documentclass[runningheads]{llncs}

\usepackage{eccv}

\usepackage{eccvabbrv}

\usepackage{graphicx}
\graphicspath{{figures/}}
\usepackage{booktabs}

\usepackage[accsupp]{axessibility}  

\usepackage{hyperref}

\usepackage{orcidlink}

\usepackage{multirow}
\usepackage{amssymb}

\begin{document}

\title{Warp-free Cross-view Geo-localization via Feature-space Consensus Mining}


\author{Zhuo Song\inst{1}\orcidlink{0009-0008-4580-0956} \and
Lian Xu\inst{2}\orcidlink{0000-0002-1759-2941} \and
Runqing Jiang\inst{1}\orcidlink{0000-0002-6603-839X} \and
Yongjian Zhang\inst{1}\orcidlink{0000-0002-6178-6532} \and
Kunhong Li\inst{1}\orcidlink{0000-0003-3297-1824} \and
Ye Zhang\inst{1}\orcidlink{0000-0002-5919-5661} \and
Yulan Guo\inst{1}\orcidlink{0000-0001-7051-841X}\thanks{Corresponding author.}
}

\authorrunning{Z. Song et al.}


\institute{
Sun Yat-sen University, Shenzhen, China \and
The University of Western Australia, Perth, Australia\\
\email{\{songz28,jiangrq3,zhangyj85,likh25\}@mail2.sysu.edu.cn, lian.xu@uwa.edu.au, zhangy2658@mail.sysu.edu.cn, guoyulan@sysu.edu.cn}}

\maketitle


\begin{abstract}

Cross-view geo-localization is challenging due to drastic viewpoint changes and large appearance discrepancies between street-level and satellite imagery. Although existing methods often use geometric warping to expose co-visible cues, such transformations rely on restrictive spatial assumptions and inevitably introduce severe visual distortions under view-dependent visibility, yielding noisy supervision and fragile correspondences. To overcome this, we propose a novel joint-view consensus-guided learning framework that entirely bypasses explicit geometric warping. Instead of forcing rigid spatial alignment, we dynamically mine and adaptively strengthen a \emph{semantic consensus} directly within the feature space. Specifically, an auxiliary joint-view pathway during training enables direct cross-view interaction, allowing each view to selectively aggregate corroborative evidence into a unified consensus representation. To resolve feature heterogeneity among the single- and joint-view streams, we introduce global pattern probes acting as a semantic dictionary to project divergent modalities into a strictly aligned metric space. 
Guided by a consensus-mediated contrastive objective, single-view embeddings are explicitly pulled toward the joint-view anchor during training, distilling this consensus-mining capability into the single-view encoders for robust retrieval at inference. 
Extensive experiments demonstrate that our method achieves state-of-the-art performance across four standard benchmarks, underscoring the importance of discovering cross-view semantic consensus for reliable geo-localization.

    
\end{abstract}





\section{Introduction}
\label{sec:intro}

\begin{figure}[!t]
\centering
\includegraphics[width=\textwidth]{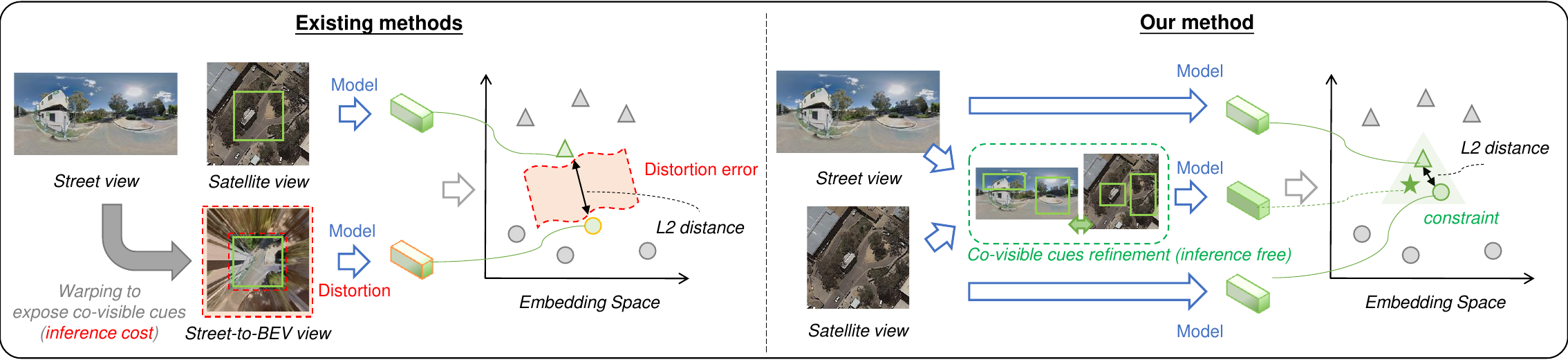}
\caption{
Comparison between warping-based methods and ours. 
Left (Existing methods): geometric warping (e.g., street-to-BEV/polar) is used to expose co-visible cues and reduce cross-view discrepancies, but it incurs extra inference cost and inevitably introduces distortion, leading to noisy supervision and fragile correspondences. 
Right (Ours): we instead refine co-visible cues in feature space by mining cross-view consensus and imposing an additional alignment constraint during training, avoiding explicit warping and its distortions; the refinement branch is discarded at inference, yielding inference-free retrieval with more robust cross-view alignment.
}
\label{fig_1:motivation}
\end{figure}

Cross-view geo-localization aims to determine the geographic location of a query image captured from ground platforms \cite{zhai2017predicting} (e.g., vehicles, handheld devices) or low-altitude UAVs \cite{ji2025game4loc} by matching it against a geo-referenced satellite image database. 
As an important alternative when GPS signals are unreliable, noisy, or denied, it has been widely adopted in real-world applications such as autonomous driving \cite{doan2019scalable}, street navigation \cite{mirowski2018learning}, object localization \cite{wilson2022object}, and augmented reality \cite{chiu2018augmented}. 
Typically, this task is formulated as a cross-view image retrieval problem \cite{hu2018cvm, cai2019ground}: the objective is to extract compact descriptors from disparate viewpoints and map them into a joint metric space, where descriptors of the same geographic location are pulled closer while those of different locations are pushed apart.
 
Establishing robust cross-view correspondences is challenging due to drastic viewpoint changes and significant appearance discrepancies between street-level and satellite imagery. 
In cross-view representation learning, \emph{co-visible content} serves as the most critical and reliable clue for matching \cite{wang2021each}. 
However, under extreme viewpoint changes, these co-visible clues exhibit severe structural discrepancies and are heavily buried within disparate, view-specific contextual noise, making them exceptionally difficult to identify and align.

To mitigate this gap, prior works often explicitly warp images (e.g., via polar transformations \cite{shi2019spatial, toker2021coming, shi2020looking} or bird's-eye-view projections \cite{ye2025cross, fervers2023uncertainty, wang2025bevsplat}) to reduce the overall geometric discrepancy between cross-view pairs. By forcibly re-parameterizing one view to geometrically resemble the other, these methods force a portion of the co-visible clues to appear with similar structural layouts, thereby easing correspondence learning. Despite their effectiveness, these geometry-driven paradigms present critical limitations (as illustrated in Fig.~\ref{fig_1:motivation}, Left). First, they strictly require prior knowledge of the cross-view imaging configuration (e.g., planar-ground assumptions) to construct the geometric priors. Second, while explicit warping highlights certain spatial layouts, it inevitably introduces severe spatial distortions and structural artifacts. This distortion not only injects noisy supervision signals into the network but also actively corrupts the discovery of other potential co-visible patterns, ultimately yielding fragile correspondences that compromise matching performance.

These observations suggest that the central challenge of cross-view geolocalization is not merely to reduce geometric discrepancies between views, but to reliably discover the shared scene evidence underlying both observations. In other words, successful matching depends on identifying the consensus between two heterogeneous views rather than forcing one view to geometrically mimic the other.
Motivated by this insight, we propose a novel joint-view consensus-guided learning framework (Fig.~\ref{fig_1:motivation}, Right) that exposes and enhances co-visible information without corrupting the original visual integrity. Instead of explicitly warping pixels, our core idea is to construct an auxiliary joint-view pathway during training to serve as a consensus-aware semantic anchor. 
Specifically, instead of treating the two views as independent streams, we enable direct cross-view interaction; this allows each view to selectively leverage corroborative cues from the other, thereby reinforcing semantically consistent content and yielding a robust cross-view consensus.
Furthermore, to address the feature heterogeneity between the single-view and joint-view streams, we introduce global pattern probes as a unified semantic dictionary. 
These probes project all streams into the exact same metric space, effectively filtering out view-exclusive noise and enforcing pattern-level cross-view consistency. 
Finally, a consensus-mediated contrastive objective explicitly pulls the single-view embeddings toward the joint-view anchor during training. Crucially, because this consensus-mining capability is intrinsically assimilated into the single-view encoders, the joint-view pathway is completely discarded at inference. 
This design achieves robust matching without any additional test-time computational overhead. \footnote{Our code is available at: \url{https://github.com/chord-sz/GeoCoM}}.

Our main contributions are summarized as follows:

\begin{itemize}
    
    \item We introduce a novel \textbf{joint-view consensus-guided learning} framework that bypasses explicit geometric warping. 
    By dynamically distilling a cross-view \emph{semantic consensus} directly in the feature space, our method provides a stable reference to guide correspondence learning without introducing geometric distortions.
    \item We design a \textbf{unified joint-view encoding} mechanism together with \textbf{global pattern probes} to mine co-visible patterns and enforce pattern-level consistency across the two views and the joint representation. 
    \item Extensive experiments on CVUSA, CVACT, University-1652, and VIGOR demonstrate that our method achieves state-of-the-art performance, and generalizes strongly under challenging cross-region settings in VIGOR.
    
\end{itemize}

\section{Related Work}

\subsection{Feature-based Cross-view Geo-localization} 
Early studies \cite{bansal2011geo, viswanathan2014vision} often relied on handcrafted descriptors to match cross-view images, but such approaches typically struggle in complex, large-scale environments. 
With the advent of deep learning, CNN-based visual representations have become a common foundation for visual matching and recognition tasks~\cite{workman2015location, guo2026deep}. 
In cross-view geo-localization, some methods directly extract global features using CNN backbones~\cite{workman2015location}, while others introduce semantic layout cues as proxies to facilitate retrieval~\cite{zhai2017predicting}. 
Song et al.~\cite{song2025unified} further leverage detailed feature representations to enrich cross-view matching. 
To obtain compact and more discriminative global descriptors, a line of work \cite{shi2020optimal, hu2018cvm, shen2023mccg} explored advanced feature aggregation techniques. 
Recently, Transformer-based architectures \cite{Zhu2023Simple, ji2025game4loc} have gained prominence for their ability to model non-local relationships. Frameworks like TransGeo \cite{zhu2022transgeo} and L2LTR \cite{yang2021cross} utilize the self-attention mechanism and \texttt{[CLS]} tokens of Vision Transformers (ViT) \cite{dosovitskiy2020image} to aggregate global context. GeoDTR \cite{zhang2023cross} further decouples image content from spatial structure, employing Transformers to learn view-dependent geometric encodings. Although these feature-construction methods improve discriminability by modeling intra-view global dependencies, their representations are constructed entirely from domain-specific, isolated inputs. Consequently, their pooling mechanisms (e.g., a single \texttt{[CLS]} token) often fail to explicitly bridge the extreme domain gap. In contrast, our approach introduces a set of globally shared pattern probes. By using the exact same probe set to aggregate dense features from both individual views and the joint-view representation via cross-attention, we explicitly project heterogeneous tokens onto a unified semantic dictionary, thereby guaranteeing structural and semantic coherence across domains.

\subsection{Geometry-based Cross-view Geo-localization}
Beyond individual feature construction, another dominant line of research emphasizes bridging the domain gap through explicit geometric alignment. 
Since ground and satellite cameras capture the world from orthogonal perspectives, these methods attempt to warp one view to geometrically resemble the other. 
SAFA \cite{shi2019spatial} analyzes the imaging characteristics of street and satellite views and applies a polar transform to warp satellite imagery toward a ground-facing view. 
LPN \cite{wang2021each} exploits correspondence via ring-partition strategies to enhance spatial consistency. 
More recently, Co-Retrieval \cite{ye2025cross} extends Sample4Geo \cite{deuser2023sample4geo} by adding a bird's-eye-view (BEV) conversion branch for the ground view, optimizing retrieval with a two-branch joint scoring mechanism.
Beyond terrestrial scenes, Chen et al.~\cite{chen2024metric} employ perspective projection to bridge rover-view and orbital imagery for lunar cross-view matching. 
Despite their effectiveness, geometry-driven alignment relies on brittle geometric assumptions and distortion-prone warping that corrupts co-visible cues and yields noisy supervision in unconstrained scenes.
In contrast, we avoid spatial warping and instead learn a training-time joint-view representation in feature space to mine and reinforce cross-view semantic consensus, yielding more robust correspondences.

\subsection{Contrastive Learning in Geo-localization}

Cross-view geo-localization is closely related to deep metric learning \cite{radford2021learning, wang2019multi}, and many studies improve retrieval by designing stronger supervised contrastive objectives \cite{varior2016gated, schroff2015facenet, vo2016localizing, chen2017beyond, cai2019ground, guo2022soft}. 
Vo et al. \cite{vo2016localizing} proposed a soft-triplet loss coupled with an exhaustive mini-batch strategy to better exploit positive and negative relations. 
Cai et al. \cite{cai2019ground} explicitly distinguished easy and hard pairs, introducing the SEH loss to emphasize informative pairs during optimization. 
Sample4Geo \cite{deuser2023sample4geo} employed symmetric InfoNCE to significantly improve optimization stability and cross-view robustness. 

However, these objectives are typically tailored for dual-view paradigms. 
To fully exploit the rich co-visible information explicitly mined by our joint-view architecture, we develop a \emph{consensus-mediated contrastive objective}. Building upon the symmetric InfoNCE loss, we augment the traditional street-to-satellite contrastive path with two crucial auxiliary branches (street-to-joint and satellite-to-joint). 
These consensus-mediated constraints regularize single-view embeddings with the joint-view anchor, encouraging them to emphasize view-shared evidence and yielding more stable and discriminative cross-view correspondences.




\section{Method}

\begin{figure}[!t]
\centering
\includegraphics[width=\textwidth]{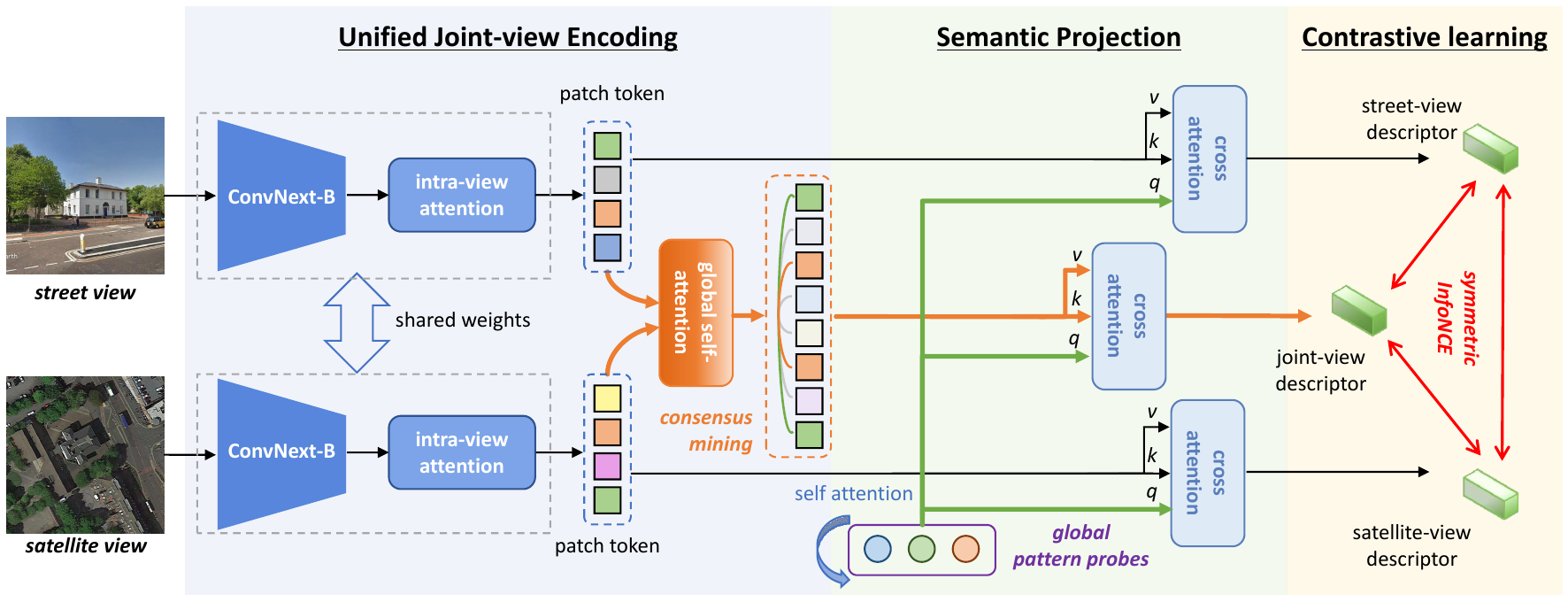}
\caption{
Overview of the proposed joint-view consensus-guided learning framework. Our pipeline consists of three key parts: (1) Unified Joint-view Encoding, where a shared backbone and an intra-view attention module extract view-specific patch tokens and a global self-attention module fuses the token sequences to mine cross-view consensus cues; (2) Semantic Projection, where a set of globally shared pattern probes (refined via self-attention) queries the street, satellite, and joint-view tokens through cross-attention to produce compact descriptors in a shared semantic basis; and (3) Contrastive Learning, where a consensus-mediated symmetric InfoNCE objective uses the joint-view descriptor as a semantic anchor to pull the single-view descriptors toward the learned consensus. During inference, the joint-view branch is removed, and retrieval is performed using only street and satellite descriptors.
}
\label{fig_2:pip}
\end{figure}

\subsection{Overview}

Cross-view geo-localization is typically formulated as a retrieval task that learns a joint metric space in which a street-view query matches its satellite counterpart from the same location. 
However, establishing reliable cross-view correspondences is challenging because the two views differ drastically in viewpoint and appearance, and each view observes only a partial, view-dependent subset of the scene. 
Under such conditions, the most trustworthy supervision comes from \emph{co-visible} evidence that can be corroborated across views, whereas view-exclusive content tends to introduce noisy gradients and lead to fragile alignments.
Motivated by this, we aim to explicitly capture and exploit cross-view \emph{consensus}—the co-visible evidence supported by both views—to regularize embedding learning and improve robustness.

As illustrated in Fig.~\ref{fig_2:pip}, we propose a joint-view consensus-guided learning framework for cross-view geo-localization. 
Given a paired street image and satellite image, we first extract dense token features with a shared backbone and perform \textbf{Unified Joint-view Encoding} (Sec.~\ref{sec:feature_extraction}) to construct a learnable joint-view representation that adaptively distills corroborative cross-view cues. 
Next, we apply \textbf{Semantic Projection via Global Pattern Probes} (Sec.~\ref{sec:feature_aggregation}), where a globally shared set of probes queries the street, satellite, and joint-view tokens via cross-attention to extract semantically consistent patterns, producing compact descriptors within a shared dictionary. 
Finally, we optimize a \textbf{Consensus-mediated Contrastive Alignment} objective (Sec.~\ref{sec:contrastive_learning}), which explicitly pulls the two single-view descriptors toward the joint-view anchor. 
Crucially, at inference time, the joint-view branch is completely removed; retrieval is performed using only the street and satellite descriptors, incurring no additional test-time computational overhead.

\subsection{Unified Joint-view Encoding for Consensus Mining}
\label{sec:feature_extraction}

Our encoding module produces token-level representations for subsequent pattern probing and contrastive alignment. 
It consists of \textbf{view-specific encoding}, which extracts structured tokens preserving the inherent layout of each input, and \textbf{joint-view encoding}, which directly bridges paired views to distill the cross-view consensus. 
While view-specific tokens capture modality-specific cues, the joint-view tokens serve as a train-time mediator that emphasizes shared semantics and suppresses view-exclusive distractors.

\textbf{View-specific Encoding.}
Given a paired street-view image $I_g$ and an aerial (satellite) image $I_a$, we first process them through a shared vision backbone to extract dense features. 
To capture long-range spatial dependencies, these features are refined by a self-attention layer, yielding two independent sets of view-specific tokens: $\mathbf{Z}_g \in \mathbb{R}^{N_g \times C}$ and $\mathbf{Z}_a \in \mathbb{R}^{N_a \times C}$. 
At this stage, $\mathbf{Z}_g$ and $\mathbf{Z}_a$ independently preserve the geometric layout and contextual cues of their respective viewpoints, without any cross-domain awareness.

\textbf{Joint-view Encoding.}
While individual embeddings capture rich view-specific details, establishing stable correspondences between them is difficult due to severe appearance gaps. 
Instead of relying on explicit and often flawed geometric warping, we propose to implicitly bridge this gap through dynamic information routing. 
By concatenating the tokens and applying global self-attention, street-view tokens can directly interact with and borrow supportive evidence from aerial-view tokens, and vice versa.

Concretely, we concatenate the view-specific tokens along the sequence dimension to form a unified token sequence $\mathbf{Z}_{concat} = [\mathbf{Z}_g; \mathbf{Z}_a] \in \mathbb{R}^{(N_g + N_a) \times C}$. 
This combined sequence is then fed into a shared Transformer encoder layer. 
Through global multi-head self-attention over $\mathbf{Z}_{concat}$, 
the network spontaneously discovers and highlights co-visible elements, performing dense cross-view relational modeling. 
The output is a highly contextualized joint-view representation $\mathbf{Z}_{joint} \in \mathbb{R}^{(N_g + N_a) \times C}$. 
By actively absorbing complementary cues from both modalities, $\mathbf{Z}_{joint}$ naturally encodes the cross-view consensus and serves as a reliable semantic anchor for subsequent alignment stages.

\subsection{Semantic Projection via Global Pattern Probes}
\label{sec:feature_aggregation}
Following the encoding stage, we obtain the street, aerial, and joint-view token sequences ($\mathbf{Z}_g$, $\mathbf{Z}_a$, and $\mathbf{Z}_{joint}$). 
A prevalent practice in Transformer-based retrieval is to aggregate these tokens using a single \texttt{[CLS]} token; however, a single global vector may under-represent the diverse, spatially distributed cues critical for precise geo-localization.
To overcome this bottleneck, we introduce a set of learnable \textbf{Global Pattern Probes} $\mathbf{P}=\{\mathbf{p}_1,\dots,\mathbf{p}_K\}$, which function as a shared dictionary of \emph{semantic bases}. 
Each probe is encouraged to capture a recurring visual primitive (e.g., boundaries, textures, layout fragments) that can manifest across viewpoints. 
In contrast to standard query-based aggregation paradigms that operate within a single domain \cite{carion2020end, ali2024boq}, our innovation lies in deploying a \emph{globally shared} set of probes across multiple highly heterogeneous streams. 
By forcing different views to be interpreted through the same set of semantic bases, the probes explicitly enforce cross-domain semantic consistency. 

Formally, we initialize $K$ learnable prototypes as the probe set $\mathbf{P} \in \mathbb{R}^{K \times C}$. 
To ensure diversity and model complex correlations among probes, we refine them through a Multi-Head Self-Attention (MSA) layer:
\begin{equation}
    \mathbf{P}' = \operatorname{MSA}(\mathbf{P}) + \mathbf{P},
\end{equation}
where $\mathbf{P}'$ represents the context-aware probes. 
Subsequently, these refined probes act as queries ($\mathbf{Q}$) in a multi-head cross-attention mechanism to scour the source tokens for semantically relevant patterns. 
We define the probing operation as:
\begin{equation}
    \operatorname{Probing}(\mathbf{Q}, \mathbf{K}, \mathbf{V}) = \operatorname{Softmax}\left(\frac{\mathbf{Q}\mathbf{K}^\top}{\sqrt{d_k}}\right)\mathbf{V}.
\end{equation}
We apply this operation in parallel to the street, satellite, and joint-view streams, using the \emph{same} probe set $\mathbf{P}'$ as queries while the token sequences serve as both keys ($\mathbf{K}$) and values ($\mathbf{V}$):
\begin{equation}
\begin{aligned}
    \mathbf{F}_{g} &= \operatorname{Probing}(\mathbf{P}', \mathbf{Z}_{g}, \mathbf{Z}_{g}), \\
    \mathbf{F}_{a} &= \operatorname{Probing}(\mathbf{P}', \mathbf{Z}_{a}, \mathbf{Z}_{a}), \\
    \mathbf{F}_{joint} &= \operatorname{Probing}(\mathbf{P}', \mathbf{Z}_{joint}, \mathbf{Z}_{joint}).
\end{aligned}
\end{equation}
The resulting feature vectors $\mathbf{F}_{*} \in \mathbb{R}^{K \times C}$ are flattened, L2-normalized, and projected to form the final descriptors $\mathbf{v}_g, \mathbf{v}_a, \mathbf{v}_{joint} \in \mathbb{R}^{D}$ (where $D = K \times C$).

The key benefit of this shared projection lies in its ability to act as an explicit information filter.
By forcing the divergent street, aerial, and joint-view tokens to be interpreted through the exact same semantic bases ($\mathbf{P}'$), GPP effectively filters out view-specific noise and enforces pattern-level cross-view consistency.
This shared-probe design distinguishes GPP from domain-specific pattern probes (SPP), which use independent probe sets for different streams.
Compared with SPP, GPP provides a common semantic basis across streams, ensuring that the resulting compact descriptors are strictly aligned in the joint metric space and laying a coherent semantic foundation for the subsequent contrastive optimization.
We empirically compare SPP and GPP in the ablation study.

\subsection{Consensus-mediated Contrastive learning}
\label{sec:contrastive_learning}
Conventional cross-view geo-localization methods typically optimize a pairwise dual-view contrastive objective (e.g., Soft-Triplet \cite{hu2018cvm} or InfoNCE \cite{radford2021learning}) between the street and satellite descriptors. 
However, establishing reliable correspondences directly between two incomplete, single-view observations is inherently fragile. 
Furthermore, the standard pairwise loss cannot leverage the rich co-visible information explicitly mined by our three-stream architecture. 
To fully exploit the constructed semantic consensus, we propose a \emph{consensus-mediated contrastive learning} paradigm. Building upon the symmetric InfoNCE loss \cite{deuser2023sample4geo}, we augment the traditional ``street-to-satellite'' contrastive path with two crucial auxiliary branches: ``street-to-joint'' and ``satellite-to-joint''.

Formally, given a mini-batch of size $B$ with embeddings $\{\mathbf{v}_g^{(i)}, \mathbf{v}_a^{(i)}, \mathbf{v}_{joint}^{(i)}\}_{i=1}^B$, the standard symmetric InfoNCE objective between the street- and satellite- views is defined as:
\begin{equation}
\label{eq:symm_infonce}
\mathcal{L}_{g \leftrightarrow a} = \mathcal{L}_{g \rightarrow a} + \mathcal{L}_{a \rightarrow g},
\end{equation}
where the uni-directional loss $\mathcal{L}_{g \rightarrow a}$ is computed as:
\begin{equation}
\label{eq:infonce_single}
\mathcal{L}_{g \rightarrow a}=-\log \frac{\exp \left(\mathbf{v}_g \cdot \mathbf{v}_{a+} / \tau\right)}{\sum_{i=1}^B \exp \left(\mathbf{v}_g \cdot \mathbf{v}_a^{(i)} / \tau\right)},
\end{equation}
where $a+$ denotes the positive aerial reference paired with the query, 
$\mathbf{v}_{a+}$ is its embedding, and $\tau$ is a learnable temperature parameter. 
The symmetric loss $\mathcal{L}_{a \rightarrow g}$ is defined analogously. 
To introduce the joint-view representation as a semantic mediator, we apply the exact same symmetric InfoNCE formulation to formulate the auxiliary losses $\mathcal{L}_{g \leftrightarrow joint}$ and $\mathcal{L}_{a \leftrightarrow joint}$. Our final synergistic overall objective jointly optimizes these three branches:
\begin{equation}
\label{eq:overall_loss}
\mathcal{L}_{total} = \mathcal{L}_{g \leftrightarrow a} + \lambda \left( \mathcal{L}_{g \leftrightarrow joint} + \mathcal{L}_{a \leftrightarrow joint} \right),
\end{equation}
where $\lambda$ is a hyperparameter balancing the primary alignment and the consensus-mediated constraints.

\section{Experiments}

\subsection{Datasets and Evaluation Metrics}

We evaluate our method on four standard cross-view geo-localization benchmarks. 
\textbf{CVUSA} \cite{zhai2017predicting} and \textbf{CVACT} \cite{liu2019lending} both provide 35,532 training and 8,884 validation/testing pairs of strictly center-aligned, one-to-one ground panoramas and satellite images. CVACT additionally includes a massive 92,802-pair test set to assess large-scale generalization.
\textbf{VIGOR} \cite{zhu2021vigor} introduces a more challenging, non-center-aligned setting comprising 90,618 aerial and 105,214 street-view images. It explicitly incorporates hard spatial distractors and features two evaluation protocols: same-area and cross-area (for testing cross-region generalization).
\textbf{University-1652} \cite{zheng2020university} is utilized to validate the versatility of our approach beyond ground-level platforms, benchmarking UAV-to-satellite matching across two specific tasks: Drone $\rightarrow$ Satellite and Satellite $\rightarrow$ Drone. 
Detailed dataset configurations and split statistics are provided in the supplementary material.

Following prior work \cite{deuser2023sample4geo, ye2025cross}, we report \textbf{Top-$k$} recall (\textbf{R@$k$}) as the primary retrieval metric: a query is counted as correct under R@$k$ if its ground-truth reference appears within the top-$k$ results.
For VIGOR, we additionally report the \textbf{hit rate} metric that tolerates a ``semi-positive'' match at top-1 under its official protocol.
For University-1652, we further report \textbf{average precision (AP)} to evaluate one-to-many and many-to-one matching scenarios.

\subsection{Implementation Details}
We adopt ConvNeXt-B \cite{liu2022convnet} as the backbone for all experiments.
To preserve higher-resolution local features, we truncate the backbone at the penultimate ConvNeXt block.
We freeze all blocks except the last one, and add a linear projection layer to reduce the channel dimension, thereby reducing the computation and memory footprint of the subsequent attention modules.
Unless otherwise stated, we use 64 global pattern probes, and study the effect of the probe number in ablations.
Following \cite{deuser2023sample4geo}, we share the weights of the backbone in the street-view and satellite branches.

We optimize the network using AdamW with a cosine decay learning-rate schedule.
For CVUSA/CVACT/VIGOR, we train for 60 epochs with batch size 64; for University-1652, we train for 10 epochs with batch size 32.
The initial learning rate is set to $2\times10^{-4}$ for all datasets, and the loss weight is fixed to $\lambda=0.5$.
Input resolutions follow standard settings: for CVUSA/CVACT, street-view and satellite images are resized to $140\times768$ and $384\times384$, respectively; for VIGOR, to $384\times768$ and $384\times384$; and for University-1652, to $384\times384$.
All models are trained on two NVIDIA GeForce RTX 3090 GPUs with PyTorch.

\begin{table}[t]
\centering
\caption{
Quantitative comparison with state-of-the-art methods on CVUSA and CVACT. ${\dagger}$ indicates polar transformation applied to aerial images, and ${\star}$ indicates BEV transformation applied to ground-view images. Best results are shown in bold.
}
\label{tab1_cvusa_cvact}
\resizebox{\linewidth}{!}{
\begin{tabular}{l|cccc|cccc|cccc} 
\hline\hline
\multirow{2}{*}{Approach}                    & \multicolumn{4}{c|}{CVUSA}                                        & \multicolumn{4}{c|}{CVACT-Val}                                    & \multicolumn{4}{c}{CVACT-Test}                                     \\
                                             & R@1            & R@5            & R@10           & R@1\%          & R@1            & R@5            & R@10           & R@1\%          & R@1            & R@5            & R@10           & R@1\%           \\ 
\hline
CVM-Net \cite{hu2018cvm}                                      & 22.47          & 49.98          & 63.18          & 93.62          & 20.15          & 45.00          & 56.87          & 87.57          & 5.41           & 14.79          & 25.63          & 54.53           \\
SAFA$^{\dagger}$ \cite{shi2019spatial}       & 89.84          & 96.93          & 98.14          & 99.64          & 81.03          & 92.80          & 94.84          & 98.17          & 55.50          & 79.94          & 85.08          & 94.49           \\
LPN$^{\dagger}$ \cite{wang2021each}        & 92.83          & 98.00          & 98.85          & 99.78          & 83.66          & 94.14          & 95.92          & 98.41          & -              & -              & -              & -               \\
TransGeo \cite{zhu2022transgeo}                                    & 94.08          & 98.36          & 99.04          & 99.77          & 84.95          & 94.14          & 95.78          & 98.37          & -              & -              & -              & -               \\
GeoDTR$^{\dagger}$ \cite{zhang2023cross}     & 95.43          & 98.86          & 99.34          & 99.86          & 86.21          & 95.44          & 96.72          & 98.77          & 64.52          & 88.59          & 91.96          & 98.74           \\
Sample4Geo \cite{deuser2023sample4geo}                                  & 98.68          & 99.68          & 99.78          & 99.87          & 90.81          & 96.74          & 97.48          & 98.77          & 71.51          & 92.42          & 94.45          & 98.70           \\
Co-retrieval$^{\star}$ \cite{ye2025cross} & 98.71          & 99.70          & 99.78          & 99.86          & 91.90          & 97.23          & 97.84          & 98.84          & 73.68          & 93.53          & 95.11          & \textbf{98.81}  \\
Ours                                         & \textbf{99.29} & \textbf{99.75} & \textbf{99.82} & \textbf{99.90} & \textbf{92.23} & \textbf{97.24} & \textbf{97.87} & \textbf{98.87} & \textbf{74.20} & \textbf{93.55} & \textbf{95.22} & 98.65           \\
\hline\hline
\end{tabular}
}
\end{table}

\subsection{Comparison with State-of-the-Art Methods}

\textit{1) Results on CVUSA and CVACT.} 
As shown in Table~\ref{tab1_cvusa_cvact}, our method establishes new state-of-the-art performance on both benchmarks. 
In particular, we achieve consistent gains in Recall@1, surpassing the second-best methods by 0.54\%, 0.33\%, and 0.52\% on CVUSA, CVACT\_val, and CVACT\_test, respectively. 
The center-aligned property of CVUSA/CVACT has motivated explicit view transformations to reduce cross-view discrepancy, such as polar-based methods (e.g., SAFA \cite{shi2019spatial}, GeoDTR \cite{zhang2023cross}) and BEV-based designs (e.g., Co-retrieval \cite{ye2025cross}). 
While these transformations can partially narrow the appearance gap under their assumptions, our approach consistently outperforms them, supporting the benefit of learning cross-view consensus in feature space via joint-view mediation and global pattern probes, rather than relying on pixel-level geometric warping.

\begin{table}[t]
\centering
\caption{Quantitative results on VIGOR. ${\dagger}$ indicates polar transformation on aerial images, and ${\star}$ indicates BEV transformation on ground views. Best results are in bold.}
\label{tab2_vigor}
\resizebox{0.9\linewidth}{!}{%
\begin{tabular}{l|ccccc|ccccc} 
\hline\hline
\multirow{2}{*}{Approach} & \multicolumn{5}{c|}{Same-area}                                                     & \multicolumn{5}{c}{Cross-area}                                                      \\
                          & R@1            & R@5            & R@10           & R@1\%          & Hit Rate       & R@1            & R@5            & R@10           & R@1\%          & Hit Rate        \\ 
\hline
SAFA$^{\dagger}$ \cite{shi2019spatial}                     & 33.93          & 58.42          & 68.12          & 98.24          & 36.87          & 8.20           & 19.59          & 26.36          & 77.61          & 8.85            \\
TransGeo \cite{zhu2022transgeo}                  & 61.48          & 87.54          & 91.88          & 99.56          & 73.09          & 18.99          & 38.24          & 46.91          & 88.94          & 21.21           \\
GeoDTR \cite{zhang2023cross}                    & 56.51          & 80.37          & 86.21          & 99.25          & 61.76          & 30.02          & 52.67          & 61.45          & 94.40          & 30.19           \\
Sample4Geo \cite{deuser2023sample4geo}                & 77.86          & 95.66          & 97.21          & 99.61          & 89.82          & 61.70          & 83.50          & 88.00          & 98.17          & 69.87           \\
Co-Retrieval$^{\star}$ \cite{ye2025cross}              & 82.18          & 97.10          & 98.17          & \textbf{99.70} & -              & 72.19          & 88.68          & 91.68          & \textbf{98.56} & -               \\
Ours                      & \textbf{83.15} & \textbf{97.43} & \textbf{98.20} & 99.59          & \textbf{94.41} & \textbf{73.60} & \textbf{91.05} & \textbf{93.45} & 98.40          & \textbf{83.01}  \\
\hline\hline
\end{tabular}
}
\end{table}


\begin{table}[t]
\centering

\begin{minipage}[t]{0.49\linewidth}
\centering
\caption{Quantitative results on University1652. Best results are in bold.}
\label{tab3_1652}
\small
\resizebox{\linewidth}{!}{%
\begin{tabular}{l|cc|cc}
\hline\hline
\multirow{2}{*}{Approach} & \multicolumn{2}{c|}{Drone2Sat} & \multicolumn{2}{c}{Sat2Drone} \\
& R@1 & AP & R@1 & AP \\
\hline
LPN \cite{wang2021each}        & 75.93          & 79.14          & 86.45          & 74.79 \\
MCCG \cite{shen2023mccg}       & 89.64          & 91.32          & 94.30          & 89.39 \\
Sample4Geo \cite{deuser2023sample4geo} & 92.65          & 93.81          & 95.14          & 91.39 \\
Camp \cite{wu2024camp}      & 94.46          & 95.38          & 96.15          & 92.72 \\
QDFL \cite{hu2025query}      & 95.00          & 95.83          & 97.15          & 94.57 \\
Ours       & \textbf{97.42} & \textbf{97.84} & \textbf{97.72} & \textbf{96.40} \\
\hline\hline
\end{tabular}%
}
\end{minipage}
\hfill
\begin{minipage}[t]{0.49\linewidth}
\centering
\caption{Ablation study on the number of learnable Global Pattern Probes (GPP). Recall@1 improves as the number of GPP increases, and saturates at 64.}
\label{tab4_ablation_q}
\small
\resizebox{\linewidth}{!}{%
\begin{tabular}{c|ccccc}
\hline\hline
\multirow{2}{*}{\begin{tabular}[c]{@{}c@{}}Number of \\ GPP (Q)\end{tabular}} & \multicolumn{5}{c}{Same-area} \\
& R@1 & R@5 & R@10 & R@1\% & Hit \\
\hline
16 & 82.24 & 97.13 & 98.10 & 99.57 & 93.68 \\
32 & 82.86 & 97.33 & 98.21 & 99.61 & 94.16 \\
64 & \textbf{83.15} & 97.43 & 98.20 & 99.59 & 94.41 \\
96 & \textbf{83.15} & \textbf{97.54} & \textbf{98.32} & \textbf{99.65} & \textbf{94.50} \\
\hline\hline
\end{tabular}%
}

\end{minipage}
\end{table}
\textit{2) Results on VIGOR.}
Table~\ref{tab2_vigor} reports results on the more challenging VIGOR benchmark. 
Our method achieves the best Recall@1/5/10 and Hit Rate under both same-area and cross-area splits. 
Notably, Co-Retrieval attains slightly higher R@1\% (Top-1\% recall), while we obtain stronger performance on the key top-$k$ recalls and Hit Rate, which better reflect practical retrieval quality. 
Since VIGOR does not satisfy the center-alignment assumption, polar-based transformations degrade more noticeably (e.g., SAFA). 
Among prior methods, Co-Retrieval is the strongest baseline: by introducing a BEV branch under a ground-plane assumption and combining it with the original street-view branch, it partially compensates for view-dependent visibility and achieves the second-best performance. 
Nevertheless, our method consistently improves over this strong competitor, especially under the cross-area split: we outperform Co-Retrieval by 1.41\% and 2.37\% in Recall@1 and Recall@5, respectively. 
These gains indicate that distilling a train-time joint-view consensus representation and aligning each single-view embedding toward this consensus provides a more robust and transferable supervision signal than branch-based geometric compensation, leading to stronger generalization under domain shift.

\textit{3) Results on University-1652.}
To verify that our method generalizes to UAV--satellite geo-localization, we evaluate on University-1652 and report results in Table~\ref{tab3_1652}.
Our method consistently achieves the best performance on both Drone$\rightarrow$Satellite and Satellite$\rightarrow$Drone tasks in terms of Recall@1 and AP. 
Compared to approaches that rely on handcrafted spatial partitioning (e.g., LPN \cite{wang2021each}), our consensus-mediated design provides a more flexible way to align cross-view evidence under large pose and scale variations. 
Moreover, compared with QDFL \cite{hu2025query}, which captures salient cues within each single-view feature map using view-specific queries, our globally shared pattern probes together with joint-view consensus alignment offer stronger and more reliable supervision, leading to higher recall and improved ranking quality (AP) in both tasks.


\begin{table}[t]
\centering
\caption{Ablation study of probe strategies and the joint-view mediator on VIGOR (same-area). SPP and GPP are mutually exclusive probe strategies. 
SPP denotes \emph{domain-specific pattern probes}, GPP denotes \emph{globally shared pattern probes}, and CM denotes the \emph{joint-view mediator} branch used for consensus-mediated alignment.}
\label{tab5_ablation_analysis}
\small
\setlength{\tabcolsep}{4pt}
\begin{tabular}{c c|c c c c c}
\hline\hline
\multicolumn{2}{c|}{Components} & \multicolumn{5}{c}{Same-area} \\
Probe Strategy & CM & R@1 & R@5 & R@10 & R@1\% & Hit \\
\midrule
-- & -- & 77.86 & 95.66 & 97.21 & 99.61 & 89.82 \\
SPP     & -- & 79.96 & 96.10 & 97.27 & 99.47 & 91.35 \\
GPP     & -- & 81.59 & 96.95 & 97.91 & 99.60 & 93.16 \\
SPP     & \checkmark & 81.96 & 97.20 & 98.15 & \textbf{99.63} & 93.76 \\
GPP     & \checkmark & \textbf{83.15} & \textbf{97.43} & \textbf{98.20} & 99.59 & \textbf{94.41} \\
\hline\hline
\end{tabular}
\vspace{-1em}
\end{table}

\subsection{Ablations}
We conduct ablation studies on the VIGOR same-area split to validate the effectiveness of our designs. 
Our ablations focus on 1) the impact of the number of pattern probes and 2) the contribution of each component.

\textit{1) Effect of the Number of Queries.} 
We investigate the sensitivity of performance to the number of pattern probes, as presented in Table~\ref{tab4_ablation_q}. 
As the probe count increases from 16 to 64, we observe a robust upward trend in retrieval accuracy, while further increasing to 96 brings no gains in Recall@1. 
Considering the trade-off between retrieval efficacy and computational overhead, we set the default number of probes to 64 for all main experiments.

\textit{2) Component Effectiveness Analysis.}
Table~\ref{tab5_ablation_analysis} analyzes two key factors in our design: the probe strategy (SPP or GPP) and the joint-view mediator (CM). Row~1 is a Sample4Geo-style baseline, where the street-view and satellite branches share the same ConvNeXt-B backbone and global descriptors are obtained via average pooling. Replacing average pooling with SPP (Row~2) improves Recall@1 from 77.86\% to 79.96\%, showing that attention-based probing with multiple learnable probes captures more diverse and discriminative cues than a single pooled vector.

Replacing SPP with GPP under the same setting without CM (Row~3) brings further gains, increasing Recall@1 to 81.59\% while consistently improving the other retrieval metrics. This indicates that a shared probe dictionary reduces view-specific projection bias, yielding cross-view descriptors that are better aligned for retrieval. Beyond the choice of probe strategy, CM also provides consistent gains in both settings: adding CM to SPP increases Recall@1 from 79.96\% to 81.96\%, while adding CM to GPP further improves it from 81.59\% to 83.15\%. Finally, the full model with both GPP and CM (Row~5) achieves the best overall retrieval performance among all variants. These results confirm that enforcing a shared pattern basis further strengthens consensus-mediated alignment, leading to more discriminative and reliable cross-view descriptors.

\begin{figure}[!t]
\centering
\includegraphics[width=\textwidth]{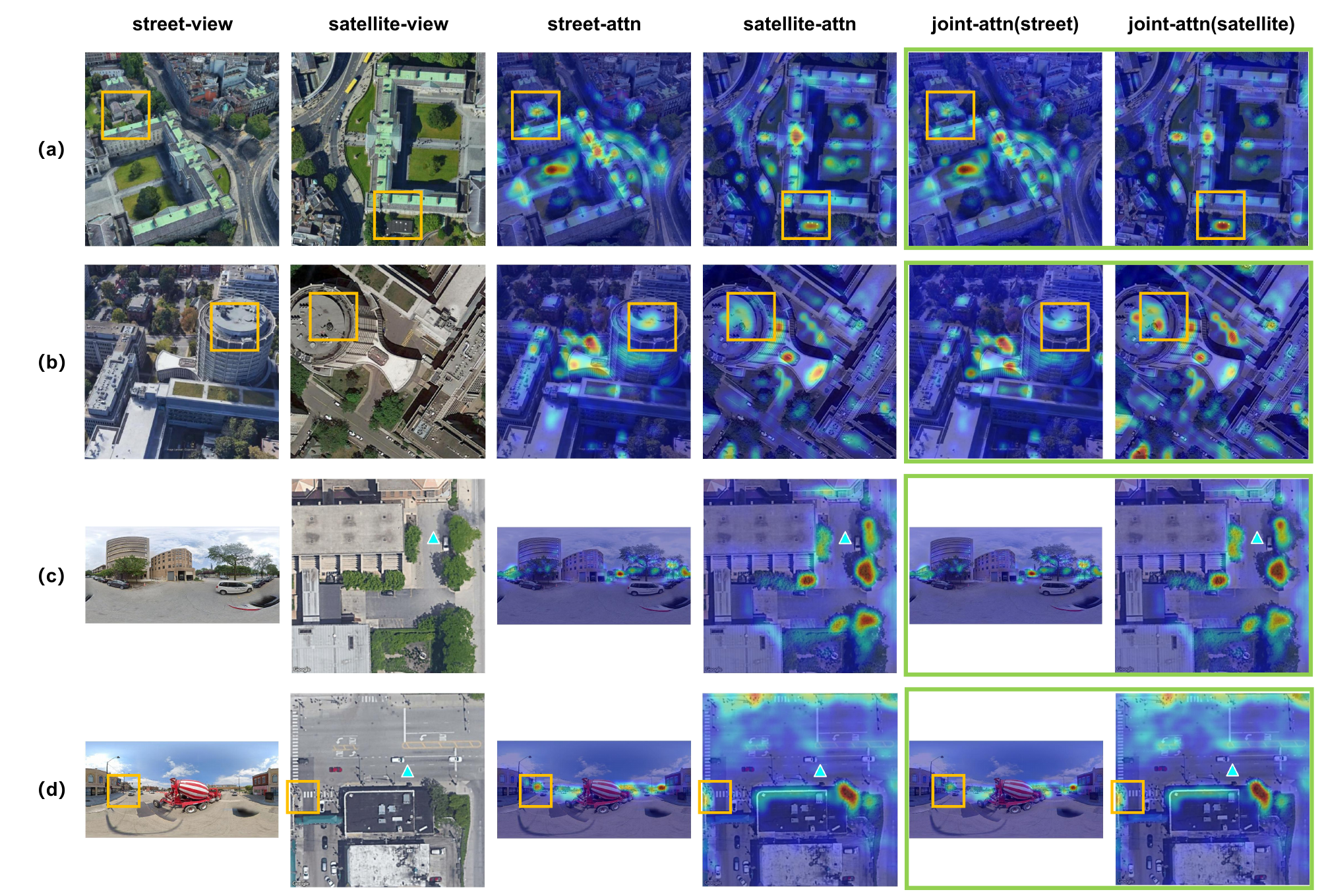}
\caption{
\textbf{Visualization of pattern-probe attention across modalities.} We visualize the cross-attention between feature maps and globally shared probes on UAV--satellite (a-b) and ground--satellite (c-d) pairs. Heatmaps with \textbf{green borders} denote the attention induced by the joint-view representation, which aligns closely with single-view inputs, visually confirming the extraction of a semantic consensus. The \textbf{cyan triangle} marks the ground-level capture location. \textbf{Orange boxes} highlight challenging, small-scale co-visible elements that our method successfully associates across views.
}
\label{fig_3:img1}
\vspace{-1em}
\end{figure}

\begin{figure}[ht]
\centering
\includegraphics[width=\textwidth]{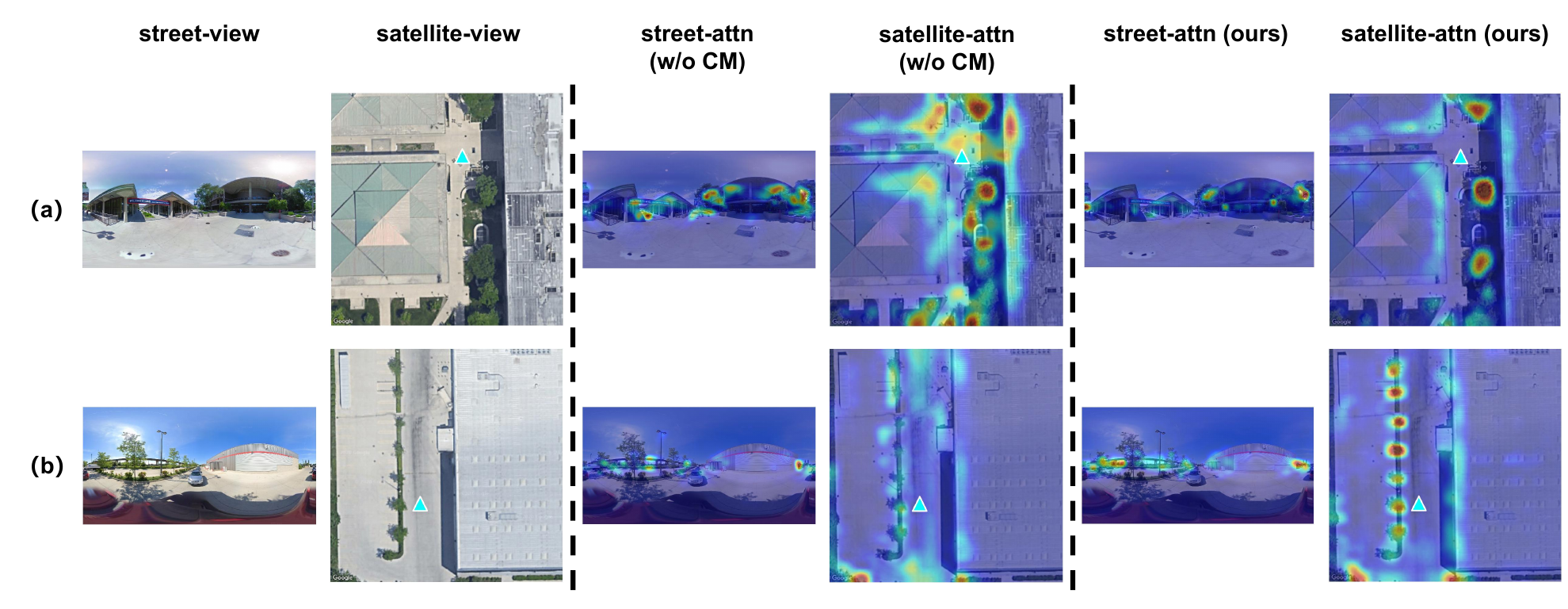}
\caption{
\textbf{Effect of consensus-mediated alignment on VIGOR.} We compare probe attention without (middle columns) and with (right columns) our consensus-mediated alignment (CM). Without this constraint, probes frequently attend to diffuse, view-exclusive areas (e.g., rooftop centers invisible from the ground). In contrast, our joint-view constraint explicitly sharpens the focus onto genuine co-visible regions (e.g., eaves and vegetation). The \textbf{cyan triangle} indicates the capture location.
}
\label{fig_4:img2}
\end{figure}

\begin{figure}[!t]
\centering
\includegraphics[width=\textwidth]{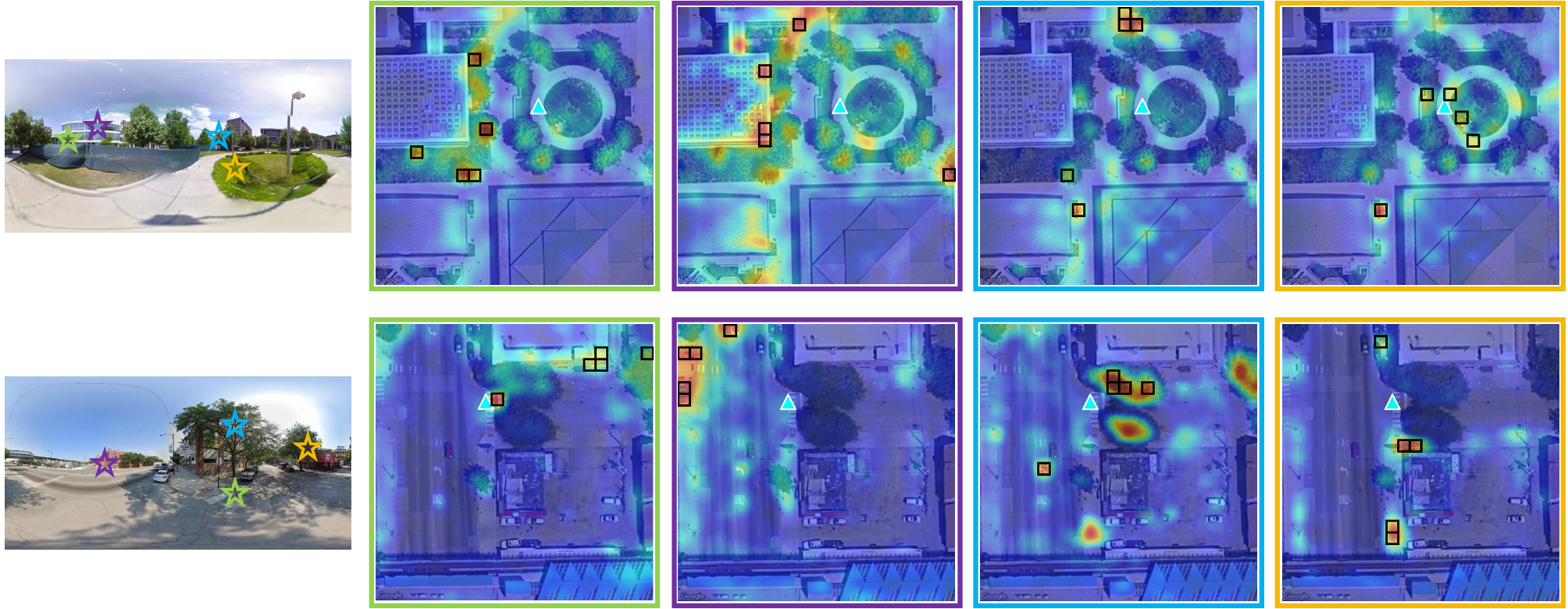}
\caption{
\textbf{Token-level inter-view interactions within the joint-view encoder.} We extract attention weights from the concatenated sequences on VIGOR to show how individual ground tokens dynamically route information from aerial tokens. \textbf{Colored stars} denote specific ground query tokens, with corresponding aerial attention heatmaps framed by matching \textbf{colored borders}. The top-5 attended aerial tokens (\textbf{black boxes}) dynamically shift to their exact semantic counterparts, demonstrating fine-grained semantic matching despite extreme viewpoint differences.
}
\label{fig_5:img3}
\vspace{-1em}
\end{figure}

\subsection{Analysis}
Our method is motivated by a key insight: under severe viewpoint changes, robust correspondence learning must be driven by \emph{co-visible} evidence corroborated across views, whereas view-exclusive content injects noisy gradients. 
To validate how our joint-view framework implicitly distills this consensus, we provide qualitative analyses focusing on (i) the semantic bases captured by global pattern probes, (ii) the regularizing effect of consensus-mediated alignment, and (iii) the token-level dynamic routing within the global self-attention module.

\textit{1) Pattern-probe Attention across Modalities.} 
To examine the discriminative cues captured by our globally shared pattern probes, we visualize the probe-to-token cross-attention on both UAV--satellite (University-1652) and ground--satellite (VIGOR) pairs in Fig.~\ref{fig_3:img1}. 
Despite extreme viewpoint disparities and frequent occlusions, the probes consistently localize co-visible, location-informative structures rather than view-specific clutter. 
In the UAV--satellite setting, attention predictably concentrates on distinctive rooftop patterns. 
Conversely, for ground--satellite pairs, the probes dynamically shift their focus to building facades, roof eaves, and surrounding vegetation. 
Notably, the attention heatmaps induced by the joint-view representation align almost perfectly with those of the single-view inputs. 
This high consistency visually proves that the probes act as a shared semantic basis, successfully filtering out domain-specific noise and extracting a unified cross-view consensus. 
Furthermore, the probes accurately associate challenging, small-scale elements across views (highlighted by orange boxes), demonstrating robust domain-invariant feature extraction.

\textit{2) Effect of Consensus-mediated Alignment.}
We further ablate the impact of our consensus-mediated contrastive objective on the challenging VIGOR benchmark (Fig.~\ref{fig_4:img2}). 
Without the joint-view constraint, the pattern probes exhibit diffuse and distracted attention, frequently responding to view-exclusive content (e.g., the center of rooftops invisible from the ground). 
Such spurious activations introduce geometric inconsistencies and degrade matching robustness. 
However, once the consensus-mediated alignment is enabled, these noisy activations are sharply suppressed. 
The network's attention becomes highly concentrated on genuine co-visible regions, such as roof edges and adjacent trees. 
This confirms that pulling individual views toward the joint-view anchor effectively compels the single-view encoders to explicitly suppress view-dependent distractors and focus on structurally stable cues.

\textit{3) Token-level Inter-view Interactions.}
To unpack the ``black box'' of the joint-view encoding branch, we analyze the global self-attention matrix within its ViT layer, which processes the concatenated ground and aerial tokens. 
By isolating the inter-view attention weights (i.e., how individual ground tokens query all aerial tokens, Fig.~\ref{fig_5:img3}), we observe a notable dynamic routing behavior. 
When querying distinct semantic entities in the ground view (e.g., a distant tree versus a nearby building facade), the high-attention regions on the aerial image dynamically shift to their exact semantic counterparts. 
This fine-grained correspondence proves that, rather than relying on rigid geometric warping, our global self-attention module allows ground tokens to actively seek and aggregate supportive evidence from the aerial view. 
This token-level interaction intrinsically bridges the domain gap, organizing cross-view features around shared semantics to boost downstream retrieval. Additional qualitative analyses are provided in the supplementary material.


\section{Conclusion}
\label{sec:conclusion}
In this paper, we propose a novel joint-view consensus-guided learning framework to tackle the severe viewpoint and appearance discrepancies in cross-view geo-localization. 
Rather than relying on rigid and error-prone geometric transformations (e.g., polar or BEV projections), we propose to  discover a cross-view semantic consensus directly within the feature space. 
By enabling direct cross-view interaction and introducing globally shared pattern probes as a unified semantic dictionary, the proposed framework distills co-visible evidence into a robust semantic anchor. 
A consensus-guided contrastive objective is further proposed to transfer this consensus knowledge to the single-view encoders, producing robust and discriminative embeddings without additional inference overhead.
Extensive experiments show that our approach achieves state-of-the-art performance across four standard benchmarks and exhibits superior cross-region generalization. 

\section*{Acknowledgements}
This work was partially supported by the National Natural Science Foundation of China (No. 62372491), the Guangdong S\&T Programme (No. 2025B0101130003), the Guangdong Basic and Applied Basic Research Foundation (2023B151512\allowbreak0087), the Science and Technology Planning Project of Key Laboratory of Advanced IntelliSense Technology, Guangdong Science and Technology Department (2023B1212060024), and the Australian Research Council grant (ARC DE260101852).


%
%
\bibliographystyle{splncs04}
\bibliography{main}
\end{document}